\documentclass[runningheads]{llncs}
\usepackage{graphicx}
\usepackage{amsmath}
\usepackage{enumerate}
\usepackage{hyperref}
\usepackage{booktabs}
\usepackage{cleveref}
\usepackage{booktabs}
\usepackage{xcolor}
\usepackage[export]{adjustbox}
\usepackage{courier}

\usepackage{caption}
\usepackage{subcaption}

\usepackage[acronym]{glossaries}
\newacronym{bert}{BERT}{Bidirectional Encoder Representations from Transformers}
\newacronym{es}{ES}{End-Systolic}
\newacronym{ed}{ED}{End-Diastolic}
\newacronym{ef}{LVEF}{Left Ventricular Ejection Fraction}
\newacronym{sv}{SV}{Systolic Volume}
\newacronym{us}{US}{Ultrasound}

\definecolor{EDLabel}{rgb}{0.75, 0.75, 0.0} 
\definecolor{ESLabel}{rgb}{0.75, 0.0, 0.75} 
\definecolor{EDPred}{rgb}{1.00, 0.00, 0.0} 
\definecolor{ESPred}{rgb}{0.0, 0.5, 0.0} 
\glsdisablehyper

\begin{document}
\title{Ultrasound Video Transformers for Cardiac Ejection Fraction Estimation}
\titlerunning{Ultrasound Video Transformers for Cardiac Ejection Fraction Estimation}
\authorrunning{Hadrien~Reynaud \emph{et al.}}

\author{Hadrien~Reynaud$^1$, Athanasios~Vlontzos$^1$, Benjamin~Hou$^1$, Arian~Beqiri$^2$, Paul~Leeson$^{2,4}$, and Bernhard~Kainz$^{1,3}$}

\institute{$^1$Department of Computing, Imperial College London, London, UK\\
$^2$Ultromics Ltd, Oxford, UK  $^3$Friedrich--Alexander University Erlangen--N\"urnberg, DE\\
$^4$John Radcliffe Hospital, Cardiovascular Clinical Research Facility, Oxford, UK
\email{hadrien.reynaud19@imperial.ac.uk}}

\maketitle 

\begin{abstract} 
Cardiac ultrasound imaging is used to diagnose various heart diseases. Common analysis pipelines involve manual processing of the video frames by expert clinicians. This suffers from intra- and inter-observer variability. We propose a novel approach to ultrasound video analysis using a transformer architecture based on a Residual Auto-Encoder Network and a BERT model adapted for token classification. This enables videos of  any length to be processed. We apply our model to the task of End-Systolic (ES) and End-Diastolic (ED) frame detection and the automated computation of the left ventricular ejection fraction. We achieve an average frame distance of 3.36 frames for the ES and 7.17 frames for the ED on videos of arbitrary length.
Our end-to-end learnable approach can estimate the ejection fraction with a MAE of 5.95 and $R^2$ of 0.52 in 0.15s per video, showing that segmentation is not the only way to predict ejection fraction. Code and models are available at \url{https://github.com/HReynaud/UVT}.

\end{abstract}

\section{Introduction}

Measurement of \gls{ef} is a commonly used tool in clinical practice to aid diagnosis of patients with heart disease and to assess options for life-prolonging therapies. The \gls{ef} is the ratio between the stroke volume, which is the difference between \gls{ed} and \gls{es} volumes, and the \gls{ed} volume of the left ventricle.

In primary and secondary care, 2D \gls{us} video acquisition of the standard apical four-chamber view is used to approximate \gls{ef} from manually delineated luminal areas in the left ventricle in one chosen \gls{ed} and one \gls{es} frame.  The biplane method of disks, which requires both 2-chamber and 4-chamber views, is the currently recommended two-dimensional method to assess \gls{ef}~\cite{folland1979assessment}, with its limitations known in literature~\cite{russo2010comparison}. The laborious nature of the data processing and substantial inter- and intra-operator variability makes this approach infeasible for high throughput studies, \emph{e.g.}, population health screening applications. Current clinical practice already neglects the recommendation to repeat this process on at least five heartbeats~\cite{lang2015recommendations} and commonly only a single measurement is acquired to mitigate clinicians' workload.  
Our objective is to estimate \gls{ef} accurately from \gls{us} video sequences of arbitrary length, containing arbitrarily many heart-beats, and to localize relevant \gls{es} and \gls{ed} frames. 

This problem has been recognized in the medical image analysis community. Initially, techniques to automatically segment the left ventricle have been proposed~\cite{8579886,9037081} to support the manual \gls{ef} estimation process.
Recently, robust step-by-step processing pipelines have been proposed to identify relevant frames, segment them, calculate the \gls{ef} and predict the risk for cardiac disease~\cite{Ouyang2020}.

To the best of our knowledge, all existing techniques for automatically processing \gls{us} video sequences follow the paradigm of discrete frame processing with limited temporal support. \gls{us} videos, however, can be of arbitrary length and the cardiac cycle varies in length. Frame-by-frame~\cite{carneiro2011segmentation,baumgartner2017sononet} processing neglects the information encoded in the change over time, or requires heuristic frame sampling methods to form a stack of selected frames to enable spatio-temporal support in deep convolutional networks~\cite{Ouyang2020}.

In this paper, we postulate that processing \gls{us} videos should be considered more similar to processing language. Thus, we seek a model that can interpret sequences across their entire temporal length, being able to reason through comparison between heartbeats and to solve tasks on the entirety of the acquired data without introducing too many difficult to generalize heuristics. 
Following this idea, we propose a new \gls{us} video \gls{es}/\gls{ed} recognition and \gls{ef} prediction network. We use an extended transformer architecture to regress simultaneously the \gls{ef} and the indices of the \gls{es} and \gls{ed} frames.

Our \textbf{contribution} is two-fold. (a) We evaluate a new paradigm for \gls{us} video analysis based on sequence-to-sequence transformer models on a large database of more than 10,000 scans~\cite{Ouyang2020} for the clinically relevant task of \gls{ef} estimation. (b) We introduce a modified transformer architecture that is able to process image sequences of variable length. 

\textbf{Related Works:} 
Early automatic detection algorithms embed videos on a manifold and perform low-dimensional clustering~\cite{Gifani_2010}. Other methods use convolutional neural networks (CNN) to locate and monitor the center of the left ventricle (LV) of the heart to extract the necessary information to categorize frames as \gls{es} or ED~\cite{zolgharni2017automatic}. In recent work, \cite{9094189} introduces a CNN followed by an RNN that utilizes Doppler \gls{us} videos to detect \gls{ed}. 

For the estimation of both the \gls{es}/\gls{ed} frame indices and the \gls{ef}, \cite{Kupinski2006} assess three algorithms performing \gls{ef} estimation in a multi-domain imaging scenario. Deep convolutional networks have been extensively used in various steps. In \cite{8579886}, the LV is segmented from standard plane US images with the help of a U-Net~\cite{ronneberger2015u}. \gls{es}, \gls{ed} frames are heuristically identified. 
In~\cite{8580137} the authors  extract spatio-temporal features directly from the input video, classifying whether the frames belong to a systole or a diastole, identifying  \gls{es} and \gls{ed} as the switching points between the two states. 
The authors of \cite{9037081} leverage deep learning techniques throughout their proposed method and directly learn to identify the \gls{es} and \gls{ed} frames which they subsequently feed into a U-Net to segment the LV followed by a metrics-based estimation of the \gls{ef}. Segmentation has been the most explored method for the analysis of cardiac functions with neural networks \cite{chen2018,hongrong2020}.
\cite{Ouyang2020} propose an end-to-end method that leverages spatio-temporal convolutions to estimate the \gls{ef}. The extracted features in conjunction with semantic segmentation of the LV enable a beat-by-beat assessment of the input.
Direct estimation of the \gls{sv} has been explored on cine MRI \cite{kong2016recognizing} and cine \gls{us}~\cite{dezaki2017deep}.
Trained to leverage temporal information, their network directly estimates the \gls{sv}, and the \gls{es}/\gls{ed} frames from single-beat cardiac videos. However, these methods require a fixed number of input frames containing a single cardiac cycle and use LSTMs for temporal feature extraction, which are well known for  forgetting initial elements when the sequences become longer~\cite{46201}.
Our approach is more closely related to the latter three approaches, which we will be comparing against. We explore the strength of transformers applied to video data like \cite{9281296,Esat2020,girdhar2019video} have done recently, and build our own architecture.

\section{Method} \label{sec:method}
\begin{figure}
    \centering
    \includegraphics[width=\textwidth,height=\textheight,keepaspectratio]{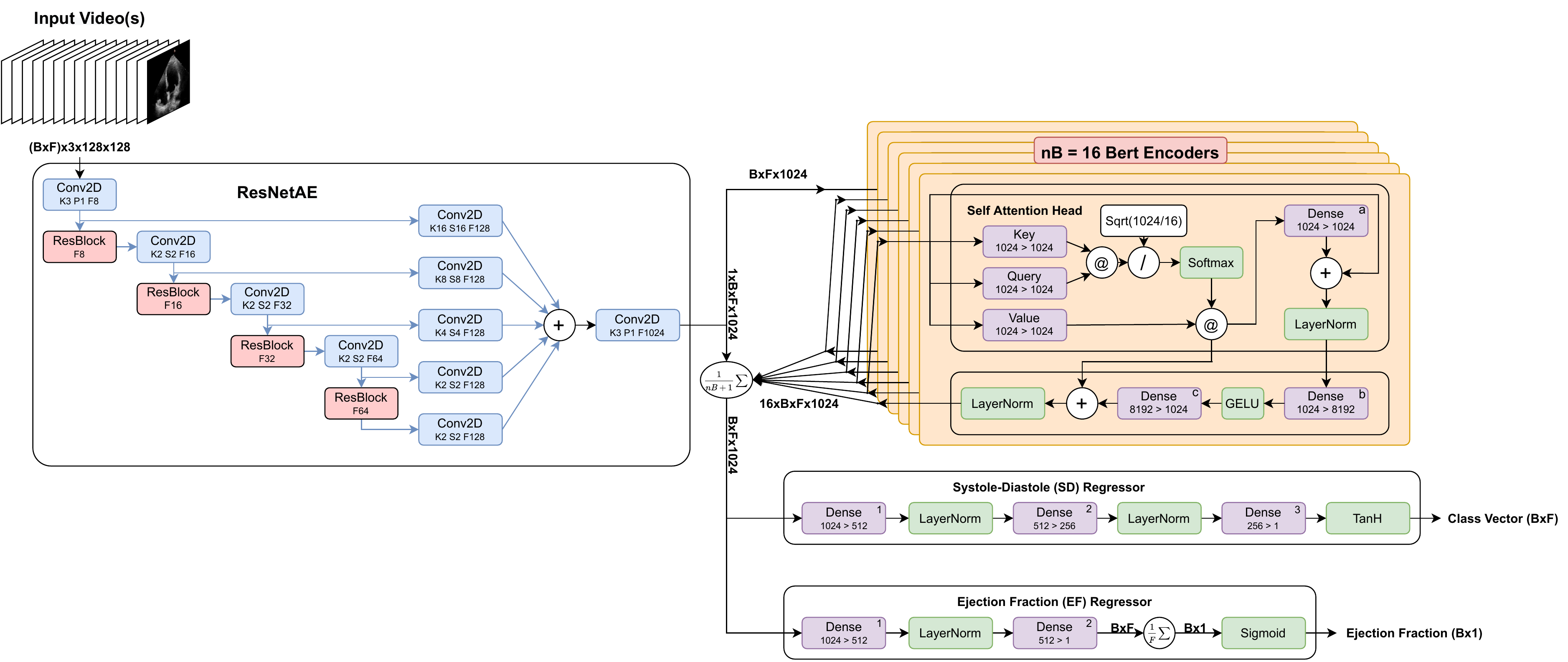}
    \caption{Overview of the proposed architecture; Left to right: Clips are reduced in dimensions through the ResAE, spatio-temporal information is extracted through the \gls{bert} and then passed to each regression branch. The Systole-Diastole (SD) Regressor predicts the \gls{es}/\gls{ed} indices while the EF Regressor predicts the \gls{ef}. The \textbf{@} operation is the dot product.}
    \label{fig:architecture}
\end{figure}

 Just as a piece of text in natural language can have variable length and is required in its entirety to be understood, a US video can be of arbitrary duration and is needed in full for accurate interpretation and reasoning. To this end, we propose to use a new transformer model to interpret \gls{us} videos.
 Our end-to-end trainable method comprises three distinct modules: (a) An encoder tasked with dimensionality reduction, (b) a \gls{bert}-based~\cite{devlin2018bert}  module providing spatio-temporal reasoning capabilities, and (c) two regressors, one labeling \gls{es} and \gls{ed} frames and the other estimating the \gls{ef}. An overview of this model is shown in Fig.~\ref{fig:architecture}. 

\noindent\textbf{Dimensionality Reduction:}
Using full size image frames directly as inputs to a transformer is infeasible, as it would require an excessive amount of computational power. To make the problem computationally tractable and to allow \gls{bert} to understand the spatial structure of the images, we make use of the encoding part of a  ResNetAE~\cite{Hou2019} to distil the US frames into a smaller dimensional embedding. 

The network uses multi-scale residual blocks~\cite{he2016identity} in both the encoder and the decoder in order to incorporate information across dimensions. We optimize the hyper-parameters associated with the AE architecture, such as the depth of the encoder, the size of the latent space, by first performing a reconstruction task on the US dataset. The encoder setup of the optimal architecture is then used as the encoding module of our method. 
Each frame from the clip is distilled by the encoder into a $1024D$ vector. The resulting embeddings are stacked together to produce the initial embedding of the clip, characterized by a shape $Batch\times N_{frames}\times 1024$.
We use these latent embeddings as the input of the transformer and the combined architecture is trained as a whole end-to-end. Cascading the encoder with the transformer model presents an important benefit as the weights of the encoder are learned to optimize the output from the transformer. This is in contrast to using a pretrained dimensionality reduction network where the encoder would be task-agnostic.

\noindent\textbf{Spatio-Temporal Reasoning:}
In order to analyze videos of arbitrary length we use a \gls{bert} encoder~\cite{wolf2019huggingface} to which we attach a regression network to build a Named Entity Recognition (NER) model for video. This acts as a spatio-temporal information extractor.  
The extracted embeddings $E$ from the \mbox{ResNetAE} encoding step, are used as inputs for the \gls{bert} encoder. As in \cite{devlin2018bert}, the $k^{th}$ encoder can be characterized as 
$\mbox{B}_k(E) = \mbox{LayerNorm} ( \mbox{D}_{k,c} ( \mbox{GELU} ( \mbox{D}_{k,b} ( \mbox{A}_k(E) ))) + \mbox{S}_k(E), \label{eq:bertenc}$
where, 
\begin{align*}
    \mbox{S}_k(E) &= \mbox{Softmax} \left( \frac{\mbox{Q}_k(E)\mbox{K}_k^T(E)}{\sqrt{\frac{\mbox{nD}}{\mbox{nB}}}} \right) \mbox{V}_k(E),   \\
    \mbox{A}_k(E) &= \mbox{LayerNorm} ( \mbox{D}_{k,a} ( \mbox{S}_k(E) ) + E ).
\end{align*}

$\mbox{S}_k(E)$ and $\mbox{A}_k(E)$ describe the Self-Attention block and Attention block respectively. The Query, Key and Value are parameterized as linear layers $\mbox{Q}_k$, $\mbox{K}_k$, $\mbox{V}_k$. While, $\mbox{D}_{k,\{a,b,c\}}$ are the intermediate linear (dense) layers in the \gls{bert} Encoder. We keep the key parameters $nB$, the number of \gls{bert} encoders, and $nD$ the dimensionality of the embeddings, similar to \cite{devlin2018bert}, setting them to $nB = 16$ and $nD=1024$ respectively. During training, dropout layers act as regularizers with a drop probability of 0.1.

\noindent\textbf{Regressing the Output:} 
Following the spatio-temporal information extraction, the resulting features from the $k$ \gls{bert} encoders $B_k(E)$ are averaged together with the ResNetAE output to 
\begin{equation}
    \mbox{M}(E) = \frac{1}{nB+1} \left( E + \sum_k^{nB} \mbox{B}_k(E) \right),
\end{equation}
and passed through two regressors tasked with predicting \gls{es}, \gls{ed} frame indices, $R_{SD}(M(E))$, and the \gls{ef}, $R_{EF}(M(E))$, respectively. We define the output of  $R_{SD}(M(E))$ as the output of three linear layers interleaved with layer-normalization, with a $tanh$ activation at the end. \gls{ef} is characterized by:

\begin{equation}
    R_{EF}(M(E)) = \mbox{Sigmoid} \left(\frac{1}{\mbox{nF}} \sum_f^{\mbox{nF}} \left( \mbox{D}\textsubscript{EF,2}\left(\mbox{LayerNorm} \left( \mbox{D}\textsubscript{EF,1} \left(\mbox{M}(E)\right)\right )\right ) \right ) \right),
\end{equation}

with $nF$ the number of input frames. 
Thus, \gls{ef} is estimated through a regression network which reduces the embedding dimension to 1 for each input frame. We then take the average of the predictions over the frames to output a single \gls{ef} prediction per video. The \gls{ef} prediction training is done with a combination of losses and regularization to address imbalance in the distribution of \gls{ef} in the training set. We use both Mean Squared Error $\mathcal{L}_{MSE}(\hat{y}, y) = \frac{1}{\mbox{nF}} \sum_f^{\mbox{nF}}  (\hat{y_f}, y_f)^2$
and Mean Average Error $\mathcal{L}_{MAE} = \frac{1}{\mbox{nF}} \sum_f^{\mbox{nF}} ||\hat{y_f}, y_f||$ 
to ensure that the network will be penalized exponentially when the error is large, but that the loss will not decrease too much when reaching small errors. Thus, the network will continue to learn even if its predictions are already close to the ground truth. A regularization term $\mathcal{R}(y) = (1-\alpha)+\left(\alpha \cdot \frac{||y-\gamma||}{\gamma}\right)$ 
helps the network to emphasize training on the \gls{ef} objective, weighing down \gls{ef} estimates which are away from $\gamma$, where $\gamma$ is chosen close to the mean of all the \gls{ef} on the training set. The $\alpha$ parameter is a scalar which adjusts the maximum amount of regularization applied to \gls{ef} close to $\gamma$. Thus, our overall objective loss can be written as
$\mathcal{L}\textsubscript{EF} = (\mathcal{L}_{MSE} + \mathcal{L}_{MAE}) \cdot \mathcal{R}(y) \label{eq:ef_loss}.$

\section{Experimentation}

\textbf{Dataset:}
For all of our experiments, we use the Echonet-Dynamic dataset~\cite{Ouyang2020} that consists of a variety of pathologies and healthy hearts. It contains 10,030 echocardiogram videos of varied length, frame rate, and image quality, all containing 4-chamber views. We pad all the frames with zeros to go from $112 \times 112$ to $128 \times 128$ pixel size frames. The videos represent at least one full heart cycle, but most of them contain three or more cycles. In each clip, however, only one cycle has the \gls{es} and \gls{ed} frames labelled. For each labelled frame, the index in the video and the corresponding ground truth segmentation of the left ventricle is available. We have access to the frame rate, left ventricular volumes in milliliters for the \gls{es} and \gls{ed} frames and the ejection fraction, computed from these heart volumes via $\mbox{LVEF} = \frac{\mbox{EDV}-\mbox{ESV}}{\mbox{ESV}}*100$.
Following Echonet~\cite{Ouyang2020}, we split the dataset in training, validation and testing subsets; dedicating $75\%$ to our training subset and equally splitting validation and testing with $12.5\%$ each.

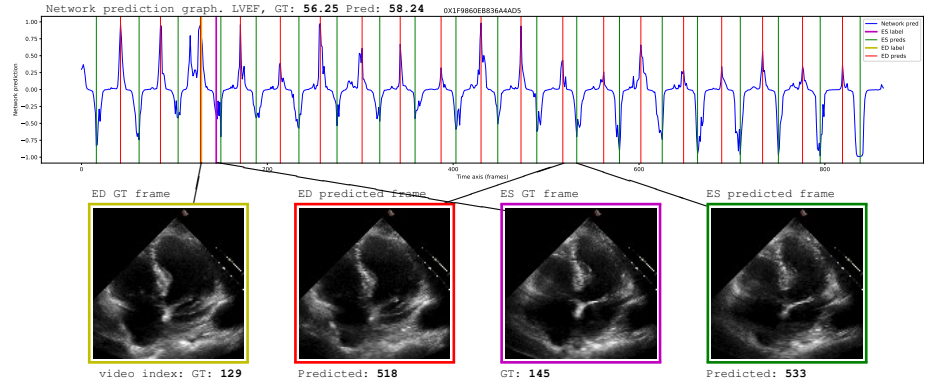
\begin{figure}[t]
\centering
\begin{subfigure}{1.0\textwidth}
    \centering
    \begin{picture}(1000,75)
    \put(0,5){\includegraphics[width=\textwidth,trim={7cm 0 7.1cm 0},clip]{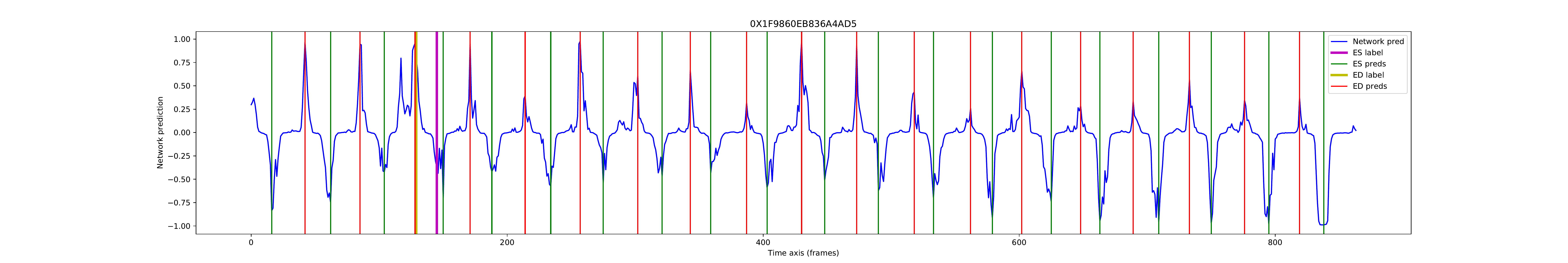}}
    \put(10,70){\scalebox{.5}{ \texttt{ Network prediction graph. LVEF, GT: \textbf{56.25} Pred: \textbf{58.24}}}}
    \end{picture}
\end{subfigure}

\begin{subfigure}{1.0\textwidth}
\centering
\begin{picture}(1000,70)

\put(70,69){\line(1,5){3}}
\put(30,10){\includegraphics[height=2.0cm,,cfbox=EDLabel 1pt 1pt]{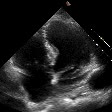}}
\put(30,3){\scalebox{.5}{ \texttt{ video index: GT: \textbf{129}} }}
\put(30,71){\scalebox{.5}{ \texttt{ED GT frame} }}

\put(165,69){\line(3,1){45}}
\put(108,10){\includegraphics[height=2.0cm,,cfbox=EDPred 1pt 1pt]{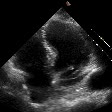}}
\put(108,3){\scalebox{.5}{ \texttt{Predicted: \textbf{518}} }}
\put(108,71){\scalebox{.5}{ \texttt{ED predicted frame}}}

\put(79,84){\line(6,-1){120}}
\put(186,10){\includegraphics[height=2.0cm,,cfbox=ESLabel 1pt 1pt]{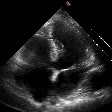}}
\put(186,3){\scalebox{.5}{\texttt{GT: \textbf{145}}}}
\put(186,71){\scalebox{.5}{\texttt{ES GT frame}}}

\put(215,84){\line(3,-1){49}}
\put(264,10){\includegraphics[height=2.0cm,,cfbox=ESPred 1pt 1pt]{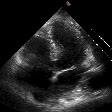}}
\put(264,3){\scalebox{.5}{\texttt{Predicted: \textbf{533}}}}
\put(264,71){\scalebox{.5}{\texttt{ES predicted frame}}}

\end{picture}

\end{subfigure}
\caption{Example network response (top), labelled pair and one predicted pair of \gls{us} frames (bottom) for a very long video. The frames are shown for the labelled ED and ES indices and one of the multiple sets of the ES/ED frames produced by the network from a different heartbeat than those of the labelled indices.}
\label{fig:results}
\end{figure}

\begin{table}[t!]
\centering

\begin{tabular}{@{}lcccccccc@{}}
\toprule
\textbf{Video sampling}  

&\multicolumn{3}{c}{\textbf{Single heartbeat}} & \multicolumn{3}{c}{\textbf{Full video}} \\ 
\midrule

&\multicolumn{6}{c}{\textit{\textbf{ED and ES index detection}}}\\
\cmidrule(lr){2-7}
& \textbf{ES} $\downarrow$  & \textbf{ED} $\downarrow$ &~~~ & \textbf{ES} $\downarrow$  & \textbf{ED} $\downarrow$  & Rejected $\downarrow$ \\
\cmidrule(lr){2-4}\cmidrule(lr){5-7}
\textbf{US \cite{dezaki2017deep}}         & 4.1                   & 3.7                   & & /                       & /                     &  /  \\
\textbf{MRI \cite{kong2016recognizing} }   & 0.44 (0.46)           & 0.38 (0.39)           & & /                       & /                     &  /  \\
\textbf{R./Cla. (ours)}                     & 1.84 (2.73)           & 2.34 (3.15)           & & \textbf{2.86 (6.43)}    & 7.88 (11.03)          &  \textbf{3}  \\ 
\textbf{R./Reg. (ours)}                     & 2.14 (2.46)           & 3.05 (3.74)           & & 3.66 (7.96)             & 9.60 (30.71)          &  4  \\ 
\textbf{M./Cla. (ours)}                     & 0.09 (1.25)           & \textbf{0.14 (1.49)}  & & 5.49 (12.94)            & 9.21 (14.15)          &  31 \\ 
\textbf{M./Reg. (ours)}                     & \textbf{0.08 (1.53)}  & 0.15 (1.89)           & & 3.35 (6.79)             & \textbf{7.17 (12.92)}  &  6  \\ 

\midrule
&\multicolumn{6}{c}{\textit{\textbf{LVEF prediction}}} \\
\cmidrule(lr){2-7}
& \textbf{MAE} $\downarrow$ & \textbf{RMSE} $\downarrow$ & \textbf{R\textsuperscript{2}} $\uparrow$ & \textbf{MAE} $\downarrow$ & \textbf{RMSE} $\downarrow$ & \textbf{R\textsuperscript{2}} $\uparrow$   \\
\cmidrule(lr){2-4}\cmidrule(lr){5-7}
\textbf{EchoNet(1)~\cite{Ouyang2020}}  & {4.22} & {5.56} & {0.79} & {4.05} & {5.32} & {0.81} \\
\cmidrule(lr){2-4}\cmidrule(lr){5-7}
\textbf{EchoNet(2)~\cite{Ouyang2020}}  & 7.35 & 9.53 & 0.40 & / & / & / \\
\textbf{R3D \cite{Ouyang2020}}      & 7.63          & 9.75          & 0.37 & /             & /             & /                      \\
\textbf{MC3 \cite{Ouyang2020}}      & 6.59          & 9.39          & 0.42 & /             & /             & /                      \\
\textbf{R. (ours)}                  & 5.54 (5.17)   & 7.57          & 0.61          & 6.77 (5.47)   & 8.70          & 0.48          \\
\textbf{M. (ours)}                  & 5.32 (4.90)   & 7.23          & 0.64          & 5.95 (5.90)   & 8.38          & 0.52          \\
\bottomrule
\end{tabular}
\caption{\textit{Top:} Results for \gls{es} and \gls{ed} detection on the test set (1024 videos) stated in \textbf{ average Frame Distance (aFD) (standard deviation)} over all frame distances. Video sampling methods are equivalent for training and testing in each experiment. The rejected column indicates the number of videos for which the network did not find clear index positions, \emph{i.e.}, not enough zero crossings in the output sequence. Our clinical expert confirmed not ideal image quality for these. US \cite{dezaki2017deep} and MRI \cite{kong2016recognizing} both train and test their approaches on single beat videos of fixed length from private datasets. 
\textit{Bottom:} Results for \gls{ef} prediction, compared to ground truth, labeled \gls{ef}. Echonet (1 \& 2), R3D and MC3 all come from \cite{Ouyang2020} and use combinations of neural networks and heuristics. "R." and "M." refer to the random and mirror video sampling methods; 
``/'' means not available, \emph{e.g.}, from literature. 
Echonet(1) is restrictive through the use of segmentations and a fixed number of frames per beat. It aggregates per-beat predictions into a final estimate for entire sequences. Processing time is 1.6s per beat and a minimum number of frames, \emph{e.g.}, 32,  have to be available for each beat during retrospective analysis.
Echonet(2) uses segmentation and all the available frames over a single, selected beat, with no further processing.
Our average processing time  is 0.15s for entire videos and our method runs in real-time  during  examination. $\uparrow$/$\downarrow$ mean higher/lower values are better.}\label{tab:ESEDDetection_table}
\end{table}

\noindent\textbf{Video sampling process:}
For training, we need videos of fixed size length to enable batch training and we need to know where the \gls{es} and \gls{ed} frames are. In our data, only single \gls{es} and \gls{ed} frames are labelled for each video, leaving many of the other true \gls{es} and \gls{ed} frames unlabeled. We choose to create 128 frames long sequences, based on the distribution of distances between consecutive \gls{es} and \gls{ed} frames in the training set. Any sequence longer than 128 frames is sub-sampled by a factor of 2. As training the transformer with unlabeled \gls{es} and \gls{ed} frames would considerably harm performance, we tried two approaches to mitigate the lack of labels:
(1) Guided Random Sampling: We sample the labeled frames and all frames in-between. We then add $10\%$ to $70\%$ of the distance between the two labelled frames before and after the sampled frames. To match the desired 128 length clip, where appropriate we pad it with black frames. The resulting clip is masked such that the attention heads ignore the empty frames. 
(2) Mirroring: We augment our clips by mirroring the transition frames between the two labelled frames, and placing them after the last labelled frame. Given the sequence 
$S = [f_{ES},f_{T_1},...,f_{T_N},f_{ED}]$, 
where $f_{T_N}$ stands for the $Nth$ transition frame, we augment the sequence such that it reaches a length superior to 128 frames by mirroring the frames around $f_{ED}$ creating the new sequence 
$S' = [f_{ES},f_{T_1},...,f_{T_N},f_{ED},f_{T_N},...,f_{T_1},f_{ES}...]$.
Once the sequences exceed 128 indices, we randomly crop them to 128 frames. Doing so ensures that the first frame is not always labelled. This augmentation follows common practices when using a transformer model in order to provide seamless computational implementation. In addition, it ensures that all the \gls{es} and \gls{ed} frames that our transformer sees, are labelled as such, while having no empty frames and retaining spatio-temporal coherence.

The labels for these frames are defined depending on the method we use to predict the frame type. When regressing, we set the \gls{es} to -1 and the \gls{ed} to 1. We adapt the heuristics used in \cite{dezaki2017deep,kong2016recognizing} to smooth the transition between the \gls{es} and \gls{ed} volumes. When using a classification objective, we define three classes: transition~(0), \gls{ed}~(1), \gls{es}~(2), with no form of smoothing.
In the mirror-sampling method we cannot use the same heuristics to smooth the volume transition when regressing. Instead, we apply an $x^3$ function scaled between the \gls{es} and \gls{ed} to soften the loss around the class peaks (-1, 1). 

\noindent\textbf{\gls{es} \& \gls{ed} Frame Detection:}
We train our end-to-end model to predict the \gls{es}, \gls{ed} indices and \gls{ef} using the two video sampling methods and two architecture variations for the SD branch. The architecture variation consists of outputting a $3 \times nF$ matrix instead of a $1 \times nF$ vector for the \gls{es}/\gls{ed} prediction and replacing the activation function by \textit{Softmax}. Instead of approximating the volume, like~\cite{dezaki2017deep,kong2016recognizing}, of the left ventricle chamber, our architecture classifies each frame in either \textit{transition}, \gls{es} or \gls{ed}. 
For the regression (Reg.) model, we use the MSE loss. For the classification (Cla.) model, we use the weighted Cross-Entropy Loss, with weights $(1, 5, 5)$, corresponding to our classes [transition, \gls{es}, \gls{ed}]. The results presented in the top half of Tab. \ref{tab:ESEDDetection_table} are obtained while training simultaneously the SD and EF branches of the network. As common in literature~\cite{dezaki2017deep,kong2016recognizing} we state average Frame Distance (aFD) as $aFD = \frac{1}{N} \sum_{n=1}^{N} \mid \hat{i}_n - i_n \mid$ and $std = sqrt{\left ( \frac{1}{N} \sum_{n=1}^{N} \mid \hat{i}_n - \Bar{i} \mid^2  \right )}$, where $i_n$ is the \gls{es} or \gls{ed} label index for the $n^{th}$ test video and $\hat{i}_n$ is the corresponding predicted index. 

\noindent\textbf{\gls{ef} Prediction:}
The prediction of the \gls{ef} is done in parallel with the \gls{es} and \gls{ed} detection. We apply a single pre-processing step which is the scaling of the value from 0-100 to 0-1. The results are presented in the 0-100 range.
For the regularization term $\mathcal{R}(y)$ we empirically chose $\alpha = 0.7$ and set $\gamma = 0.65$ to match the over-represented \gls{ef} values from the \gls{ef} distribution in the training set. Results for \gls{ef} are summarized in the bottom half of Tab. \ref{tab:ESEDDetection_table} and compared to results from literature. 

\noindent\textbf{Ablation Study:}
We tested the capacity of our M. Reg. model. From Tab. \ref{tab:ablation} we observe that using only four \gls{bert} encoders achieves results similar to the original architecture. Removing the EF or SD branch has little to no impact on the other branch. Most of the computing time is used in the ResNetAE encoder while most of the parameters are in the \gls{bert} encoders.

\begin{table}[t]
\centering
\begin{tabular}{@{}lcccc|c|c|cc@{}}
\toprule
Exp         & \gls{bert}s   & Em.      & Linear   & Seq. len.   & \gls{es}$\mid$\gls{ed}        &  MAE$\mid$RMSE$\mid$R\textsuperscript{2} & Inference  &  Params.\\
\midrule
Ours        & 16            & 1024          & 8192          & 128           & 3.35$\mid$\phantom{0}7.17     & 5.95$\mid$8.38$\mid$0.52                  & 0.15s     &  346.8M\\
Reduced 1   & 4             & 256           & 1024          & 64            & 4.28$\mid$10.12               & 6.06$\mid$8.37$\mid$0.51                  & 0.13s     &  \phantom{00}6.8M\\
Reduced 2   & 1             & 128           & 512           & 64            & 8.18$\mid$35.42               & 7.34$\mid$9.96$\mid$0.30                  & 0.13s     &  \phantom{00}2.7M\\
No EF       & 16            & 1024          & 8192          & 128           & 3.71$\mid$\phantom{0}7.24     & /                                         & 0.15s     &  346.3M\\
No SD       & 16            & 1024          & 8192          & 128           & /                             & 5.95$\mid$8.38$\mid$0.52                  & 0.15s     &  346.2M\\
\bottomrule
\end{tabular}
\caption{Ablation study. \gls{es}$\mid$\gls{ed} columns show the aFD and MAE$\mid$RMSE$\mid$R\textsuperscript{2} shows \gls{ef} scores. (Em. = Embeddings; Seq. len. = Sequence length)}
\label{tab:ablation}
\end{table}

\noindent\textbf{Discussion: }
Manual measurement of \gls{ef} exhibits an inter-observer variability of $7.6\%-13.9\%$~\cite{Ouyang2020}. From Tab.~\ref{tab:ESEDDetection_table}, we can speculate that transformer models are en par with human experts for \gls{ef} estimation and \gls{ed}/\gls{es} identification.

We observe that our model predicts on average a 3\% higher \gls{ef} than measured by the manual ground truth observers, especially on long videos. We hypothesize that the model learned to correlate spatio-temporal image features to ventricular blood volume, therefore sometimes selecting different heartbeats than labelled in the ground truth, \emph{e.g.}, Fig.~\ref{fig:results}. While guidelines state that operators should select the cycle with the largest change between the \gls{ed} area and the \gls{es} area, in practice, operators will usually just pick a ``good'' heart cycle, which might also be the case for our ground truth. This hypothesis requires clinical studies and cross-modal patient studies in future work. 

A limitation of our study is that aFD is not an ideal metric, since over-prediction would lead to inflated scores. Our model predicts distinct \gls{es}/\gls{ed} frames as shown in Fig.~\ref{fig:results}. This can also be shown by calculating the average heart rate from peak predictions. When accounting for frame rate differences this results in a reasonable 53 beats per minutes.

\noindent\textbf{Implementation: } PyTorch 1.7.1+cu110 with two Nvidia Titan RTX GPUs.

\section{Conclusion}
We have discussed probably the first transformer architecture that is able to analyze \gls{us} videos of arbitrary length. Our model outperforms all existing techniques when evaluated on adult cardiac 4-chamber view sequences, where the task is to accurately find \gls{es} and \gls{ed} frame indices. We also outperform heuristic-free methods on \gls{ef} prediction. In the future we would expect transformers to play a more prominent role for temporally resolved medical imaging data.\\

\noindent\textbf{Acknowledgements:}
This work was supported by the UKRI Centre for Doctoral Training in Artificial Intelligence for Healthcare (EP/S023283/1) and Ultromics Ltd.

\bibliographystyle{splncs04}
\bibliography{reference}

\end{document}